\documentclass{IEEEtran}
\usepackage{graphicx} 
\usepackage{amsmath}
\usepackage{bm}
\usepackage{amsthm}
\usepackage{subfigure}
\usepackage{url}
\newtheorem*{theorem-non}{Theorem}
\usepackage{balance}
\usepackage[flushleft]{threeparttable}
\usepackage{xspace}
\newcommand*{\eg}{e.g.\@\xspace}
\newcommand*{\ie}{i.e.\@\xspace}
\usepackage{multirow,makecell}
\usepackage{tikz}
\def\checkmark{\tikz\fill[scale=0.3](0,.35) -- (.25,0) -- (1,.7) -- (.25,.15) -- cycle;} 
\makeatletter
\newcommand*{\etc}{%
	\@ifnextchar{.}%
	{etc}%
	{etc.\@\xspace}%
}
\makeatother

\begin{document}
\title{A Dilated Inception Network for \\ Visual Saliency Prediction}
%
%
%

\author{
	Sheng~Yang,
	Guosheng~Lin,
	Qiuping~Jiang,~\IEEEmembership{Member,~IEEE,}
	Weisi~Lin,~\IEEEmembership{Fellow,~IEEE,}
	\thanks{S. Yang, G. Lin, W. Lin are with the School of Computer Science and Engineering, Nanyang Technological University, Singapore 639798. (e-mail: syang014@e.ntu.edu.sg; gslin@ntu.edu.sg; wslin@ntu.edu.sg)}
	\thanks{Q. Jiang is with the Faculty of Information Science and Engineering,
		Ningbo University, Ningbo 315211, China. (email: jiangqiuping@nbu.edu.cn)}
}

%



\maketitle

\begin{abstract}

Recently, with the advent of deep convolutional neural networks (DCNN), the improvements in visual saliency prediction research are impressive. One possible direction to approach the next improvement is to fully characterize the multi-scale saliency-influential factors with a computationally-friendly module in DCNN architectures. In this work, we proposed an end-to-end dilated inception network (DINet) for visual saliency prediction. It captures multi-scale contextual features effectively with very limited extra parameters. Instead of utilizing parallel standard convolutions with different kernel sizes as the existing inception module, our proposed dilated inception module (DIM) uses parallel dilated convolutions with different dilation rates which can significantly reduce the computation load while enriching the diversity of receptive fields in feature maps. Moreover, the performance of our saliency model is further improved by using a set of linear normalization-based probability distribution distance metrics as loss functions. As such, we can formulate saliency prediction as a probability distribution prediction task for global saliency inference instead of a typical pixel-wise regression problem. Experimental results on several challenging saliency benchmark datasets demonstrate that our DINet with proposed loss functions can achieve state-of-the-art performance with shorter inference time.

\end{abstract}


%
\IEEEpeerreviewmaketitle
\section{Introduction}
%
%
%
%

\IEEEPARstart{V}{isual} attention mechanism refers to the ability of Human Vision System (HVS) to automatically select the most salient or interested regions from natural scenes by filtering out redundant and unimportant visual information for further processing. Around $10^{8}$-$10^{9}$ bits per second of visual data enters into our eyes as reported in \cite{borji2013state}. Without the help of visual attention mechanism, the HVS is impossible to handle and process this large volume of data in real-time. Therefore, it is important to understand and simulate the behavior of visual attention to advance a wide range of visual-oriented multimedia applications such as image retrieval \cite{gao2015database}, image retargeting \cite{lin2013patch}, video summarization \cite{evangelopoulos2013multimodal}, image and video compression \cite{li2017closed,hadizadeh2014saliency}, visual quality assessment \cite{gu2016saliency,jiang2018optimizing,kim2015transition}, object detection \cite{ye2017salient,borji2014salient,Liu_2018_CVPR}, virtual reality content design \cite{sitzmann2018saliency}, and more.


In general, visual attention is stimulated by two types of factors: bottom-up and top-down. Bottom-up saliency-driven attention, which is derived directly from the distinctiveness of visual stimuli, helps people to rapidly focus on conspicuous points/regions automatically. In contrast, top-down attention is task-driven and usually can help people to deal with specific visual tasks. 

This paper focuses on modeling the task-free bottom-up visual attention mechanism by predicting human eye fixations on natural images. The study of this visual attention modeling, commonly referred as visual saliency prediction/detection, is an active problem in the field of computer vision and neuroscience. Typically, a saliency map, where a pixel with brighter intensity indicates a higher probability of attracting human attention, is generated as the output of the developed visual saliency detection models.

Most of classic bottom-up saliency prediction models \cite{itti1998model,harel2007graph,zhang2013saliency} are biologically inspired. They mainly adopt multiple low-level hand-crafted features, such as intensity, color, and so on, and combine these features in a heuristics way (\eg center-surround contrast \cite{itti1998model}, graph-based random walk \cite{harel2007graph}, \etc.). However, these low-level hand-crafted features and their heuristics combination are insufficient to represent the wide variety of factors that contribute to visual saliency \cite{kruthiventi2017deepfix,liu2016deep,cornia2016predicting}. 


With the advent of Deep Convolutional Neural Networks (DCNN), the feature extraction and combination could be formulated in a data-driven manner through fully end-to-end training. With such techniques, researchers are able to get rid of finding more discriminant hand-crafted features and designing more powerful feature combination methods for further improving the cutting edge of the saliency prediction task. At present, DCNN-based saliency models have defeated the classical saliency prediction models in all challenging saliency datasets \cite{jiang2015salicon,judd2009learning,Judd_2012}. Within these DCNN-based models, the use of multi-scale contextual features \cite{kruthiventi2017deepfix,liu2016deep,dodge2017visual,wang2017deep}, which aims to characterize the diverse saliency-influential factors at different receptive field sizes, makes them stand out. However, state-of-the-art saliency models suffer from the huge computation cost by fully exploiting these comprehensive feature representations.

In this work, we propose a DCNN architecture called Dilated Inception Network (DINet) for bottom-up visual saliency prediction. In order to fully exploit the multi-scale contextual features, an efficient yet effective dilated inception module (DIM) is involved. The original inception module \cite{szegedy2015going} utilizes multiple convolutional layers with different kernel sizes to serve as multi-scale feature extractors with various receptive fields. In contrast, our DIM uses parallel dilated convolutions with different dilation rates \cite{chen2016deeplab} to capture more comprehensive and effective multi-scale contextual features with much less computation cost.

In addition, it has been reported that the saliency prediction task can be formulated as a probability distribution prediction problem \cite{jetley2016end}. Existing softmax normalization-based probability distribution distance metrics outperform the commonly-used regression loss functions by utilizing this formulation. We further propose a set of linear normalization-based probability distribution distance metrics to train our model by replacing the softmax normalization with our linear regularization on the loss function. As demonstrated in the experiments, the saliency prediction model trained with our loss functions achieves better performance than the same architecture trained with either softmax normalization-based loss functions or standard regression loss functions.   

%

%

From the network architecture perspective, the proposed DINet can be decomposed into two parts: encoder and decoder networks. The DCNN-based backbone network is paired with our DIM to serve as the encoder. Then, the encoded features are forwarded to a simple yet effective fully convolutional decoder network for saliency inference. The whole encoder-decoder model is trained end-to-end by using a set of linear normalization-based probability distribution distance metrics as loss functions. Experimental results validate that our DINet with proposed loss functions achieves state-of-the-art performance on the various saliency benchmark datasets in terms of both efficiency and efficacy. The source code of the DINet and its pre-trained model is publicly available\footnote{https://github.com/ysyscool/DINet}.

In summary, our main contributions are threefold:
\begin{itemize}
    \item  We propose an efficient and effective dilated inception module (DIM) to capture the multi-scale contextual features. The scale diversity is enriched by introducing paralleled dilated convolutions with various dilation ratios at lower computation cost.

	\item A set of linear normalization-based probability distribution distance metrics are proposed as loss functions to optimize our DINet. They provide an additional linear regularization leading to a promising performance gain.

	\item The computation cost is further reduced by replacing the deconvolution layers with a fully convolutional decoder structure. As a result, the whole model is efficient to achieve real-time performance.
	
	
\end{itemize}

The rest of this paper is organized as follows. The related works on visual saliency prediction are summarized in Section \ref{sec:related0}. The proposed DINet and optimization method are illustrated in Section \ref{sec:approach}. The detail analysis and the peer comparison on public benchmarks will be provided in Section \ref{sec:exp}, and the conclusion is given in Section \ref{sec:con}.

\section{Related Work}
\label{sec:related0}
%
%
%
%

In this section, we first review the previous saliency prediction models with deep learning architectures. Then, we particularly summarize the existing deep saliency models with multi-scale feature extraction module. 
\subsection{Deep Learning-Based Visual Saliency Prediction}
\label{sec:related}


Nowadays, the advances in deep learning have already boosted the progress in saliency prediction. To the best of our knowledge, the first attempt to use convolutional neural networks to predict visual saliency was introduced by Vig \emph{et al.} in 2014 \cite{vig2014large}. Their model, called eDN, consists of three individual and different shallow networks (from 1 layer to 3 layers) for feature extraction. However, this model is inferior to some traditional unsupervised saliency models \cite{harel2007graph,zhang2013saliency} mainly due to the limited depth of their networks. After that, researchers seek to use deeper models (\eg AlexNet \cite{krizhevsky2012imagenet} in \cite{kummerer2014deep,huang2015salicon}, VGGNet \cite{simonyan2014very} in \cite{kruthiventi2017deepfix,wang2017deep}, and ResNet \cite{he2016deep} in \cite{liu2016deep,cornia2016predicting}) and utilize the fully convolutional network (FCN) \cite{long2015fully} framework for fully leveraging the powerful capabilities of DCNN models in contextual feature extraction.


Conventionally, current DCNN models utilize some down-sampling operations (\eg max pooling and convolutions with strides) to reduce the computation cost and enlarge the receptive field in their subsequent layers. We denote the ratio of the input image spatial resolution to the output resolution by \emph{output\_stride}, and we use this term to simplify the later descriptions. The more usage of down-sampling operations, the higher \emph{output\_stride} is. Note that, higher \emph{output\_stride} means the feature maps in the top layers have a relatively smaller spatial resolution. Such limited spatial information cannot support effective dense prediction of saliency \cite{liu2016deep,cornia2016predicting}. A naive approach, presented in ML-Net \cite{cornia2016deep} and MxSalNet \cite{dodge2017visual}, to increase the spatial resolution in top layers is simply removing some down-sampling operations in some of the layers. This indeed increases the spatial resolution, but has an undesirable side effect that negates the benefits: removing down-sampling correspondingly reduces the receptive field size in subsequent layers. Since the size of the receptive field affects the amount of contextual information which is essential to the final saliency inference, such reduction in receptive field size is suboptimal. Therefore, a trade-off between the spatial resolution of feature maps and the computation cost should be guaranteed while maintaining suitable receptive field sizes. With such considerations, several state-of-the-art deep saliency prediction models \cite{kruthiventi2017deepfix,liu2016deep,cornia2016predicting} adopt dilated convolution \cite{yu2015multi,chen2016deeplab,yu2017dilated} strategy to increase the receptive field sizes of the top layers, compensating for the reduction in receptive field size induced by removing down-sampling operations.

\begin{table*}[]
	\centering
	\scriptsize
	\begin{threeparttable}
	\caption{
		Overall comparison of recent deep saliency prediction models.
	}
	\label{Saliencys}
	{		
		\begin{tabular}{|c|c|c|c|c|c|c|c|}
			\hline
			Model      & Backbone network & \emph{output\_stride} & Input/Inputs & Multi-scale & Loss function                  & pixel/PD & Center-bias  \\ \hline \hline
			SALICON \cite{huang2015salicon}    & AlexNet/VGG16/GoogleNet            & 16             & multi inputs & yes (IPN)         & KLD                            & PD       & implicit             \\ \hline
			DeepGazeII \cite{kummerer2016deepgaze}& VGG19            & 16             & single input & no          & BCE (softmax)             & PD       & explicit           \\ \hline
			PDP  \cite{jetley2016end}      & VGGNet           &  N/A          & single input & no          & statistical distances (softmax) & PD       & implicit           \\ \hline
			ML-Net    \cite{cornia2016deep} & VGG16            & 8              & single input & yes (Skip-layer)         & Euclidean distance                 & pixel    & explicit         \\ \hline
			SAM       \cite{cornia2016predicting} & VGG16/ResNet50   & 8              & single input & no          & KLD + NSS                        & PD + pixel & explicit         \\ \hline
			SALGAN     \cite{pan2017salgan} & VGG16            & 16             & single input & no          & BCE                            & pixel    & implicit           \\ \hline
			DeepFix   \cite{kruthiventi2017deepfix} & VGG16            & 8              & single input & yes (Inception)         & Euclidean distance                 & pixel    & explicit          \\ \hline
			DSCLRCN   \cite{liu2016deep}  & VGG16/ResNet50 + Places-CNN   & 8              & multi inputs & yes (Skip-layer)        & NSS                            & pixel    & implicit            \\ \hline
			DVA     \cite{wang2017deep}   & VGG16            & 16             & single input & yes (Skip-layer)        & BCE                            & pixel    & implicit            \\ \hline
			MxSalNet \cite{dodge2017visual}  & VGG16            & 8              & single input & yes (Skip-layer)        & Euclidean distance + CCE           & pixel    & explicit          \\ \hline \hline
			DINet (Ours)       & ResNet50         & 8              & single input & yes (Inception)          & statistical distances (linear)  & PD       & implicit            \\ \hline
		\end{tabular}
	}
    \begin{tablenotes}
	\small
	\item 	KLD: Kullback-Leibler divergence, PD: probability distribution, BCE: binary cross entropy,  N/A: not available, NSS: normalized scanpath saliency, CCE: categorical cross entropy. 
	\end{tablenotes}
	\end{threeparttable}
	\vspace{-4mm}
\end{table*}

As demonstrated by previous studies \cite{kruthiventi2017deepfix,wang2017deep}, multi-scale contextual features are essential to the visual saliency prediction problem. In fact, the foundation for this conclusion is from the intuition that visual information is processed at various scales by human eyes \cite{szegedy2015going,yan2013hierarchical}. As for loss function, most of the existing DCNN-based saliency models directly use the typical pixel-wise classification or regression loss functions whereas saliency prediction is evaluated on the whole saliency maps. In \cite{jetley2016end}, Jetley \emph{et al.} propose to use loss functions based on statistical distances with softmax normalization for training saliency models. Their results demonstrate the improvement by considering saliency maps as probability distributions.

Regarding the center-bias, which is a well-known phenomenon in human vision, some of the saliency models learn the center-bias explicitly by their designed modules, such as the location biased convolutional layer in DeepFix \cite{kruthiventi2017deepfix}. However, with the help of large-scale dataset--SALICON (Saliency in Context) \cite{jiang2015salicon}, DCNN-based saliency models can learn this bias implicitly and solely from the training data \cite{jetley2016end,he2018catches}.



Table \ref{Saliencys} provides a comparison of recent deep saliency models and our proposed model. The models with multi-scale inputs will integrate multi-scale contextual features while some models with single input still can capture these due to their architectures, as detailed in the next section. For a specific saliency model, whether it is a pixel-level regression or probability distribution prediction model depends on the loss function used in the training stage. Our experimental results also verified that probability distribution prediction models perform better than the pixel-level regression ones under the same architecture.

%


\subsection{Existing Multi-Scale Feature Extraction Deep Learning Architectures}

\begin{figure}[h]
	\centering
	\includegraphics[page=1,trim = 5mm 2.5mm 5mm 5mm, clip, width=1.0\linewidth]{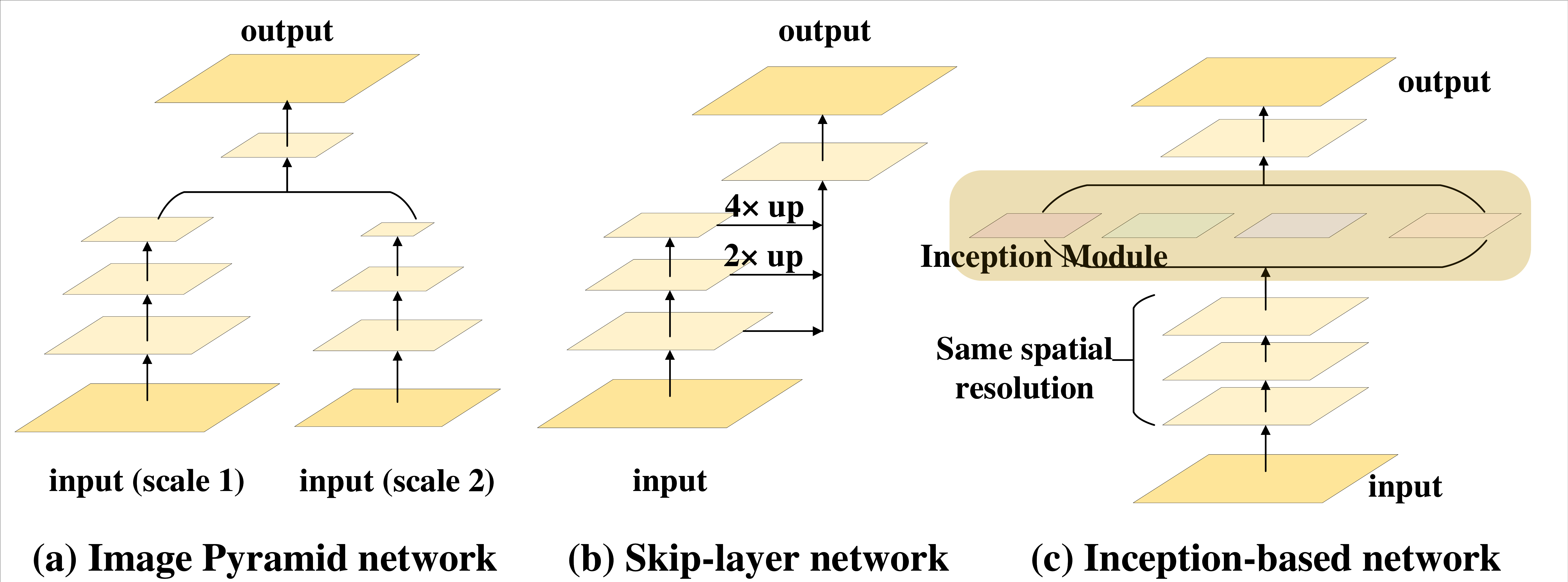}
	\caption{The illustration of existing deep learning architectures to capture multi-scale information in saliency prediction.}
	\label{fig:arch11}
	\vspace{-4mm}
\end{figure} 

In Fig. \ref{fig:arch11}, we illustrate the existing deep architectures aiming at capturing multi-scale contextual features in saliency prediction. These models can be roughly classified into three categories: i) Image Pyramid Network; ii) Skip-layer Network; and iii) Inception based Network.

\subsubsection{Image Pyramid Network}
The most straightforward way to learn multi-scale feature representations can be found in \cite{huang2015salicon,liu2015predicting}. Their idea is to apply duplicate or multiple feature extractor networks with the multi-scale inputs, as shown in Fig. \ref{fig:arch11}(a). The backbone networks in this image pyramid network (IPN) are parallel and may have different structures, corresponding to multiple scales. The outputs of these parallel networks are merged and fed into the following decoder network to generate the final saliency map. Such architectures with multi-scale inputs indeed can learn the multi-scale contextual features. Nevertheless, training and testing these models are not economic in term of computation cost and memory usage. 

\subsubsection{Skip-layer Network}
Due to the down-sampling operations in the common backbone networks, the output of each convolutional blocks is usually in different spatial resolution. The first several convolutional blocks learn the low-level image features while the features learned from the deeper blocks will contain semantic information and discriminative pattern with various receptive fields \cite{zeiler2014visualizing}. Based on this principle, architectures with skip-layers have been proposed in\cite{wang2017deep,kummerer2014deep,cornia2016deep}. Instead of applying duplicate parallel encoder networks on multiple-scale inputs, skip-layer network captures multi-scale contextual features by concatenating the outputs of different layers with increasingly larger receptive fields and \emph{output\_stride}, as illustrated in Fig. \ref{fig:arch11}(b). It is obvious that skip-layer network is more efficient than the IPN-based model. Furthermore, the skip-layer network can efficiently utilize intermediate features while the conventional way only utilizes the topmost features. Despite the high efficiency, a main problem in the skip-layer network is that spatial information gradually reduced in the higher layers due to the double-edged effect of down-sampling operations. Direct up-sampling and concatenating these feature maps from different layers will bring uncertainty and ambiguity into the saliency inference and restrict its further improvement by incorporating multi-scale features. 

\subsubsection{Inception-based Network}

As demonstrated in Fig. \ref{fig:arch11}(c), inception-based network, as introduced in the DeepFix model \cite{kruthiventi2017deepfix}, avoid the previous problem by utilizing the dilated convolutions and removing some down-sampling operations in the backbone network. Therefore, its output still has sufficient spatial information to support the dense prediction. Inception modules, proposed in the well-known GoogleNet \cite{szegedy2015going}, are attached to the top of the backbone network to capture multi-scale contextual features. The main idea of inception module is to use convolutions with multiple kernel sizes. However, existing inception module is not very economic in both computation and optimization. Our work is based on this type of network. Specifically, we revise the original inception module to have more powerful multi-scale feature extraction capacity in a computationally-friendly manner, as will be presented in Section \ref{sec:DI}. It should be noted that, in addition to the GoogleNet \cite{szegedy2015going}, our dilated inception module also take the advantage of the atrous spatial pyramid pooling (ASPP) module in the DeepLab model \cite{chen2016deeplab}, which has succeeded in semantic segmentation. We apply those parallel dilated convolutional layers to form our dilated inception module and thus obtain the state-of-the-art performance in saliency prediction.
%


\section{Our approach}
\label{sec:approach}
\begin{figure}[h]
	\centering
	\includegraphics[page=3,trim = 5mm 3mm 5mm 5mm, clip,width=1.0\linewidth]{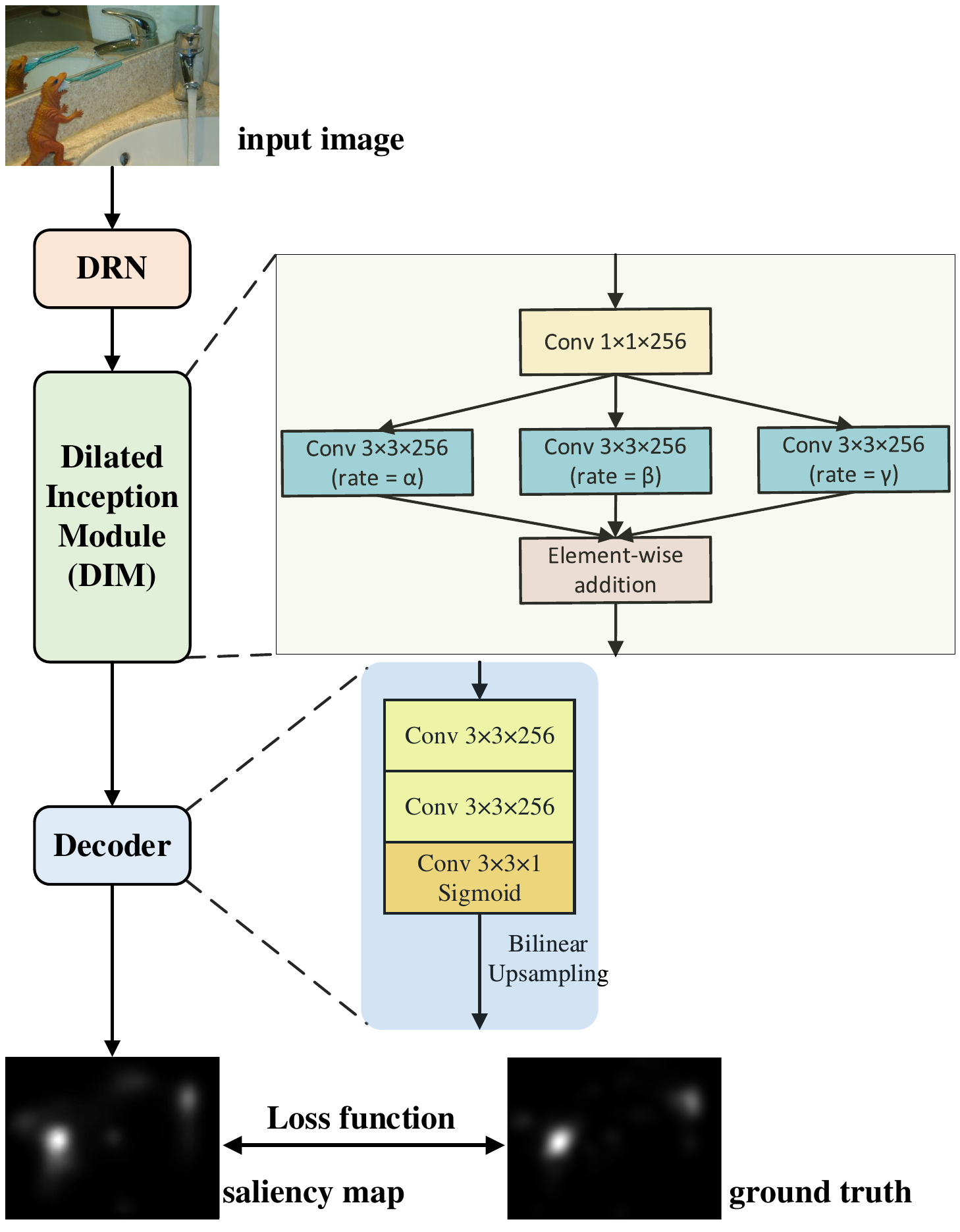}
	\caption{The architecture of our proposed DINet saliency prediction model.}
	\label{fig:model}
\end{figure}

In this section, we illustrate the architecture of our DCNN-based saliency prediction model--DINet (Dilated Inception Network). The whole model is depicted in Fig.\ref{fig:model}. Our model starts from the Dilated Residual Network (DRN) \cite{yu2017dilated} which is used as the primary feature extractor to extract dense feature maps with relatively larger spatial resolution. We propose to attach an effective dilated inception module to the top of DRN for capturing the multi-scale features. A simple yet effective decoder network is employed at the end for converting these features into the saliency maps. Furthermore, since the saliency map can be viewed as a probability distribution, we propose a set of linear normalization-based probability distribution distance metrics for training our DINet to better measure the gaps between our saliency predictions and ground-truths. 


\subsection{Dilated Convolution and Dilated Residual Network}
\label{sec:DC}
\subsubsection{Dilated Convolution}
The main idea of dilated convolution is to insert “holes”(zeros) in convolutional kernels to increase the receptive field, thus enabling dense feature extraction in DCNN. Since the usage of dilated convolutions is the core of our model, we simply revisit its concept and properties here.

\begin{figure}[h]
	\centering
	\scriptsize
	\includegraphics[page=2,trim = 5mm 2mm 5mm 5mm, clip, width=1.0\linewidth]{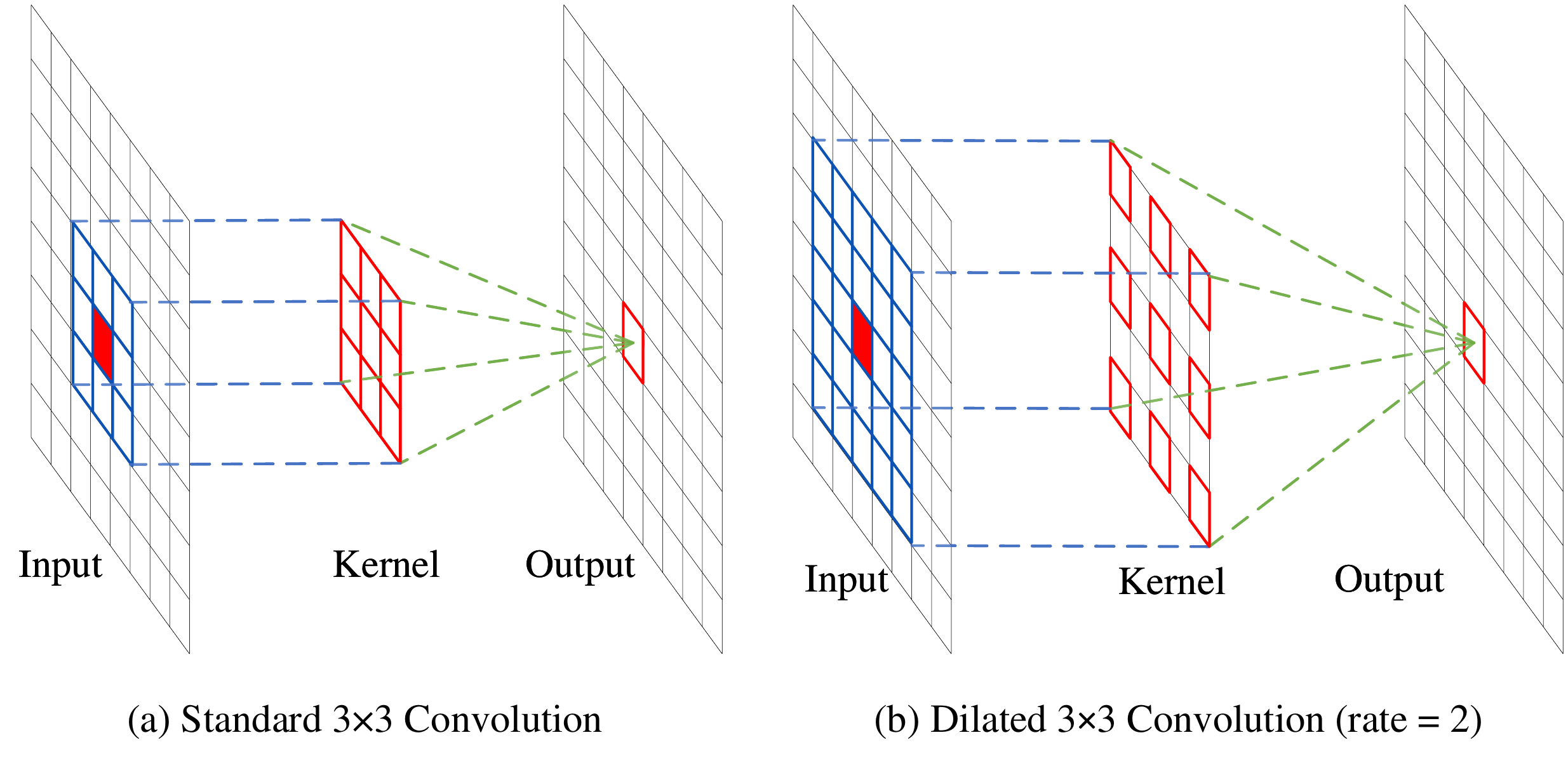}
	\caption{A comparison between standard convolution (a) and dilated convolution (b).}
	\label{fig:dc1}
\end{figure}

In general, for each spatial location $i$, dilated convolution is defined as:
\begin{equation}
\label{eq:dc}
y[i] = \sum_l  x[i+r\cdot l]w[l], 
\end{equation}
where $y[i]$ and  $x[i]$ denote the output and input on location $i$, respectively. $w$ is the convolutional filter and $r$ is the dilation rate to sample the input. Dilated convolution is implemented by inserting $r-1$ zeros between two consecutive spatial positions in the original filter $w$ along each spatial dimension. For a $k \times k $ convolutional kernel, the actual size of the dilated convolutional kernel is $k_d \times k_d $, where $k_d = k+ (k-1) \cdot (r-1)$. It should be noted that dilated convolution still only have $k \times k $ meaningful kernel parameter. The standard convolution is a special case of dilated convolution with $r = 1$. A comparison between standard convolution and dilated convolution is illustrated in Fig.\ref{fig:dc1}.
It is obvious that a dilated $3\times3$ convolutional kernel with $r= 2$ sample the feature maps like a $5\times5$ standard convolutional kernel, which means the receptive field of the outputs after these two kernels is roughly the same. With this observation, we can arbitrarily change the field-of-view of dilated convolutional kernels via choosing different dilation rate. By incorporating dilated convolutions into the encoder network, the dilated encoder network is capable of preserving the spatial resolution and compensate the receptive field reduction/shrinkage caused by removing some pooling or stride convolutional layers in the original encoder network.

\subsubsection{Dilated Residual Network}

There are two commonly used pre-trained backbone networks for saliency prediction: VGG-16 and ResNet-50. In addition, both of these two backbone networks have their corresponding dilated versions. Thanks to the residual learning introduced by He et al. in \cite{he2016deep}, the ResNet can be trained very deeply for more comprehensive feature extraction. Existing works also support that (dilated/plain) ResNet-50 based saliency models perform better than those based on (dilated/plain) VGG-16. In this work, we employ the commonly used ResNet-50 as our backbone network.

ResNet-50 backbone network has five blocks of convolutional layers. The \emph{output\_stride} of the plain ResNet-50 network is 32 which will lead to some ambiguities in dense predictions. In dilated ResNet-50 \cite{liu2016deep,cornia2016predicting}, to obtain relatively larger spatial resolution without too much computation cost increase, the original three convolutional blocks are kept fixed while the Conv4 and Conv5 blocks are modified by removing down-sampling operations and replacing the standard convolutions inside these blocks by dilated convolutions with dilation rate of 2 and 4, respectively. As a result, the \emph{output\_stride} of dilated ResNet-50 is 8 which results in a good compromise between the spatial resolution and computation cost.

\subsection{Decoder Network}


In our framework, the DRN acts as a basic encoder network. Note that a decoder network is needed to generate the saliency map from the encoded features in DRN. One conventional decoder network is built by stacking deconvolutional layers which can also help in up-sampling the coarse feature maps into dense ones. However, up-sampling these non-dense feature maps by deconvolutions inevitably need extra heavy computations and also bring some non-smoothing patterns inside them \cite{odena2016deconvolution}. Thanks to the DRN backbone network, the encoded feature maps have relatively denser spatial information. Therefore, the deconvolutional layers are no longer used in our decoder network. 

Instead, our decoder network is very simple since it only consists of three stacked standard convolutional layers with one bilinear up-sampling operation in the end. This number of convolutional layers is determined by our experiments in Section \ref{sec:dec}. The first two layers have 256 $3\times3$ convolutional kernels with the ReLU activation. The last convolutional layer is the prediction layer. It has only one $3\times3$ convolutional kernel with the sigmoid activation to generate the down-sampled version of the prediction. The reason for using sigmoid activation function in this layer is related to the range of saliency value where each pixel belongs to [0,1]. The outputs can be rescaled into this target interval by this activation. After these three convolutions, the resolution of the outputs is still lower than the inputs since no up-sampling operations are involved. To reduce the computation cost and also keep a relatively good performance, a bilinear up-sampling operation is applied in the end. Compared to the existing efforts, our decoder network is simple yet effective. The baseline model for this paper is the combination of DRN and this decoder network. In fact, we insert a $2048\times 1 \times 1 \times 256$ convolutional layer between the DRN and decoder network to reduce the number of parameters in this baseline model. To our surprise, the performance of our baseline model has no visible change with such modification. Generally, for each branch in the inception module, it has a $1 \times 1$ convolutional block at the beginning for the same purpose. When constructing the DINet, we replace this newly inserted layer by a dilated inception module, as presented in the next section.

%

\subsection{Proposed Dilated Inception Module}
\label{sec:DI}

\begin{figure}[h]
	\centering
	\scriptsize
	\includegraphics[page=4,trim = 5mm 3mm 5mm 5mm, clip,width=1.0\linewidth]{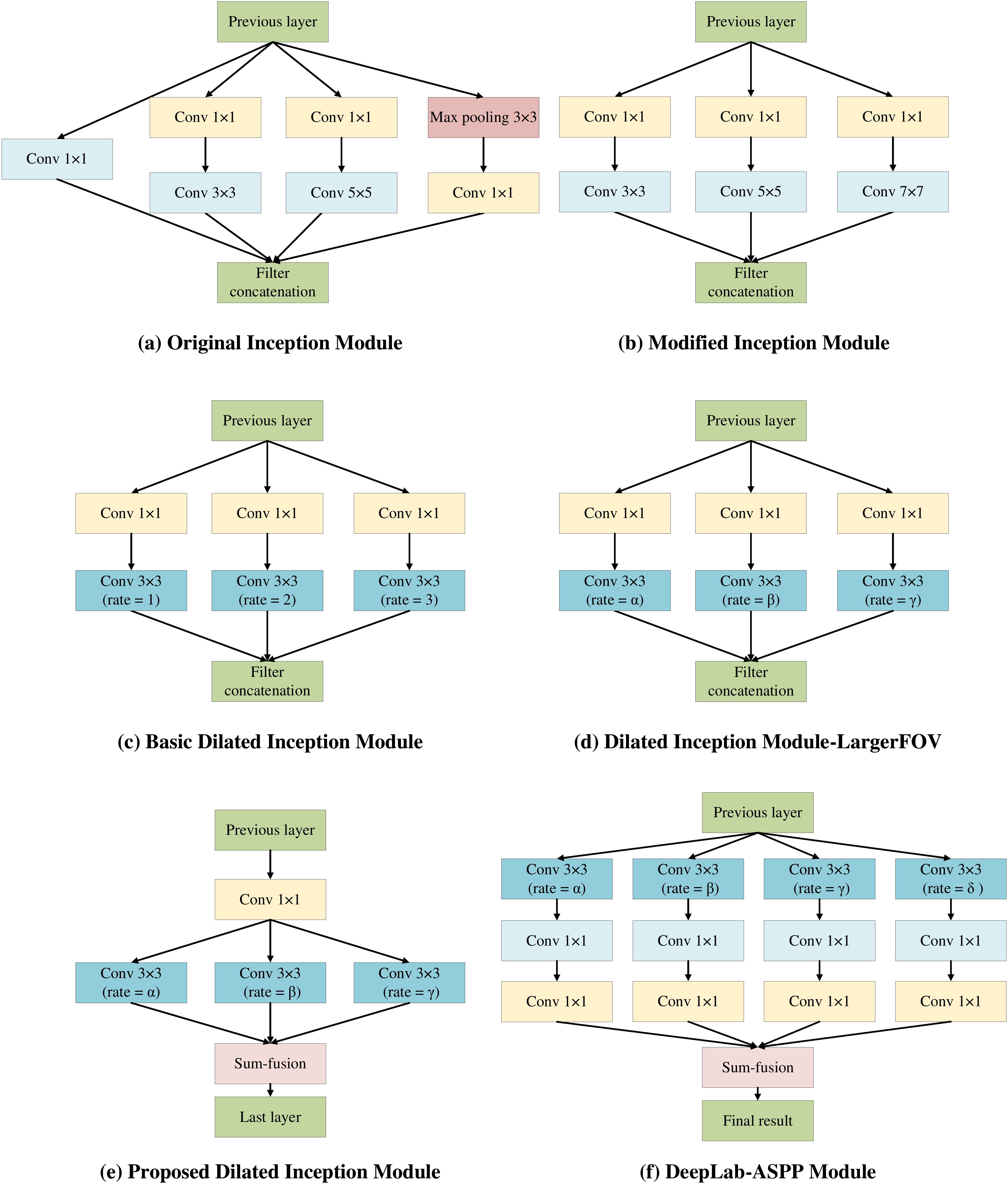}
	\caption{		
		The inception module with its variations and the ASPP module. Module (a) is the original inception module \cite{szegedy2015going}. Module (b), (c), and (d) are three variants. Module (e) is our final proposed dilated inception module (DIM). Module (f) is the DeepLab-ASPP module \cite{chen2016deeplab}. The yellow $1 \times 1$ convolutional blocks have the ability of dimensionality reduction. 
	}
	\label{fig:inc}
	\vspace{-4mm}
\end{figure} 

\begin{table*}[]
	\scriptsize
	\centering
	\caption{
			The comparison of the baseline model and other models with different multi-scale	context feature extraction modules. The model (Baseline+ Inception(e)) in bold is our final proposed DINet model.
	}
	\label{inception1}{		
		
		\begin{tabular}{|l|c|c|c|c|}
			\hline
			Model                                 & Total  \#params        & Extra params (\%)  & Best validation loss & Average Inference time \\ \hline
			Baseline (DRN + Decoder)              & 25.27M                  & 0                      & 0.2776               &        72.40s                                  \\ \hline
			ResNet + Skip-layer + Decoder           & \multirow{2}{*}{26.84M} & \multirow{2}{*}{6.21}  & 0.2793               &         48.52s                                 \\ \cline{1-1} \cline{4-5} 
			Baseline + Skip-layer                 &                         &                        & 0.2739               &          76.38s                                \\ \hline
			Baseline + IPN                        & 26.38M                  & 4.39                   & 0.2732               &          100.54s                                \\ \hline
			Baseline + Inception(a)               & 30.84M                  & 22.04                  & 0.2720                &                 89.50s                         \\ \hline
			Baseline + Inception(a) - 1$\times$1 branch & 29.72M                  & 17.61                  & 0.2721               &       87.91s                                   \\ \hline
			Baseline + Inception(b)               & 32.94M                  & 30.35                  & 0.2701               &    90.22s                                      \\ \hline
			Baseline + Inception(c)               & \multirow{2}{*}{29.27M} & \multirow{2}{*}{15.83} & 0.2696               &  81.08s                                        \\ \cline{1-1} \cline{4-5} 
			Baseline + Inception(d)               &                         &                        & 0.2673               &      81.22s                                   \\ \hline
			\textbf{Baseline + Inception(e)}      & 27.04M                  & 7.00                      & 0.2679               &        77.39s                                  \\ \hline
			DRN + ASPP-S                          & \multirow{2}{*}{42.70M} & \multirow{2}{*}{67.98} &     0.2679
			&      116.92s                                     \\ \cline{1-1} \cline{4-5} 
			DRN + ASPP-L                          &                         &                        &  0.2684
			&    141.38s                                     \\ \hline
		\end{tabular}		
	}
	\vspace{-4mm}
\end{table*}

The proposed module is derived from the inception module which intends to capture the multi-scale contextual information from the inputs \cite{szegedy2015going}. The principal idea of the original inception module is to utilize multiple convolutional layers with different kernel sizes working as multi-scale feature extractors with various receptive field sizes, as shown in Fig.\ref{fig:inc}(a). Unlike the well-known GoogLeNet \cite{szegedy2015going} which is stacked by several customized inception modules with carefully designed topologies, inception module acts as a single plug-in module in our model to diversify the receptive fields of those encoded features from the output of DRN. \par


To get rid of designing hyper-parameters for each branch as GoogLeNet, the filter numbers in each branch are all fixed to 256 in our experiments. By inserting the inception module between the DRN and decoder network, the performance of our new model is improved obviously with acceptable extra parameters and computations. However, we find that the branch of $1 \times 1$ convolutional block has limited influence on final results. The reason for this phenomenon is related to the definition of saliency since it reflects the distinctiveness of each image location with respect to their surroundings. $1 \times 1$ convolution focus on the location itself and cannot obtain the necessary spatial information from its neighbors. In the original inception module, the max-pooling branch is added by following the convention. To further investigate and explore the convolutional layers within inception module, we replace this max-pooling branch by one $7 \times 7$ convolutional layer after one $1 \times 1$ convolutional layer, as shown in Fig.\ref{fig:inc}(b). With the help of $7 \times 7$ convolutional block, the modified inception module can extract more diverse and wider field-of-view (FOV) features. For simplification, we denote the parameters number of a $256\times1 \times 1\times256$ convolutional layer (without bias term) as $W$. Therefore, $7 \times 7$ convolutional layer in inception module (b) has $7^2 W =49 W$ parameters to be determined, which is much more than $5 \times 5$ convolution ($25W$ parameters) and $3 \times 3$ convolution ($9W$ parameters). The total number of parameters in the modified inception model needs an additional $32W$ parameters compared to the original inception model, which result in larger computation cost and longer inference time. \par

Recall the dilated convolutions introduced in section \ref{sec:DC}, dilated convolutions can be used to replace the large kernel standard convolutions under the same receptive field, as shown in Fig.\ref{fig:inc}(c). $7 \times 7$ and $5 \times 5$ convolutions in the modified inception module can be replaced by $3 \times 3$ dilated convolutions with dilation rate of 3 and 2, respectively. After this replacement, the dilated inception module can perform the similar or even better results as the modified inception module with $(7^2 + 5^2 - 2\times3^2)W = 56W$ parameters less. In DRN, dilated convolutions are used in a cascaded way to preserve the spatial resolution and compensate the reduction in receptive fields. While in dilated inception module, dilated convolutions are used in a parallel way to enhance the encoded features with diverse and comprehensive field-of-views.\par

Furthermore, the dilation rate of these three parallel dilated convolutions can be arbitrarily changed, as denoted by $[\alpha,\beta,\gamma]$. Considering that the last convolutional block of the DRN has set the dilation rate equal to 4, our dilated inception module can be viewed as an extended convolutional block of the DRN with a combination of three parallel dilated convolutions inside. In our experiments, we set $[\alpha,\beta,\gamma] = [4,8,16]$ which show a great improvement from the primary dilated or original inception module. The receptive fields of the outputs after our dilated inception module are diverse and relatively large which contribute to incorporate various contextual information at different scales. This module with larger FOV is depicted in Fig.\ref{fig:inc}(d). We further reduce the computational complexity of our model by building a bottleneck type of dilated inception module (DIM), as shown in Fig.\ref{fig:inc}(e). On the one hand, we use one single $1 \times 1$ convolutional layer in the top to replace the existing individual ones in the different branches for dimensionality reduction. On the other hand, the filter concatenation is replaced by sum-fusion (element-wise addition) which can also help in dimensionality reduction and efficient computation. As a result, our final dilated inception module only brings an additional $27W$ parameters which indicate only three extra $3 \times 3$ convolutional layers are added, compared to the baseline model. Furthermore, with the help of this computationally-friendly module, our proposed DINet can reach more than 50 FPS inference time for input images of size $240 \times 320$. 


In the literature, the atrous spatial pyramid pooling (ASPP) module \cite{chen2016deeplab} also utilize parallel dilated convolutions for learning multi-scale feature representations, as shown in Fig.\ref{fig:inc}(f). In this module, the features extracted at different dilation rates are further processed in separate branches and sum-fused to generate the final results. In contrast, our DIM is just a single plug-in module and its outputs are still features, rather than the final results. Since these two modules share the same idea of using the parallel dilated convolutions, it is also reasonable to use ASPP module to replace our DIM and its followed decoder network for saliency prediction. Directly insert this ASPP module on the top of DRN cannot guarantee that every pixel in the final results is in the range of [0,1]. We add an extra linear scaling operation after sum-fusion to solve this. ASPP module has two variants: ASPP-S and ASPP-L. The only difference in these two is the setting of dilation rates. ASPP-S has smaller dilation rates ($[\alpha,\beta,\gamma,\theta] = [2,4,8,12]$) while ASPP-L has larger rates ($[6,12,18,24]$). The information of these two ASPP-based saliency models is reported in the last two rows in Table \ref{inception1}. As observed from this table, with the help of huge extra parameters, model (DRN + ASPP-S) can obtain a similar performance to our DINet. Compared to the ASPP module, our DIM only need one decoder network to generate the saliency predictions since we have the sum-fusion before the decoder rather than after it. Another reason for longer inference time in the ASPP-based model is that our DIM performs the $1 \times 1$ convolution before the dilated convolutions for dimension reduction while ASPP directly uses dilated convolutions to process these features from DRN. Specifically, the difference between the dilated convolutions part of ASPP and our DIM in \#parameters is $8\times3^2\times4=288W$ versus $8+3^2\times3=35W$.

\par
Besides, we also investigate other existing multi-scale context feature extraction frameworks, such as image pyramid network (IPN) with shared backbone network and skip-layer network, into our baseline model. The overall comparison among these models is listed in Table \ref{inception1}. Extra params (\%) term indicates the percentage of the number of additional parameters involved when using this model compared to the baseline model. The best validation loss term means that the smallest loss results of the models on SALICON validation dataset \cite{jiang2015salicon}. The loss function used in here is the linear normalization-based total variation distance, as discussed in the next section. The detailed evaluation results corresponding to these loss values are reported in Table \ref{table:maa}. Average inference time term is the average time of these models for predicting 5,000 validation images with 5 repeats under the same experimental conditions. Among these models in Table \ref{inception1}, our DINet achieves a relatively good trade-off between the validation performance and inference speed.

\subsection{Loss Function}

\label{sec:Loss}
Most saliency models directly predict saliency maps via optimizing loss functions designed for pixel-wise regression/classification. However, saliency map can be viewed as a probability distribution of human fixations over the whole image \cite{jetley2016end}. Pixel-wise prediction, where each pixel is predicted individually, may suffer from the global inconsistency problem as it ignores the inter-pixel relationship. Therefore, it is reasonable to use off-the-shelf probability distribution distance metrics as loss functions. In order to convert the predicted saliency map and its corresponding ground-truth into probability distributions, a normalization method should be applied first. Here, we improve the existing method \cite{jetley2016end} by replacing their softmax normalization with a simple linear regularization.

Base on the validation experimental results, we select the total variation distance as the loss function. Besides, the unnormalized version of total variation distance is the $\ell1$-norm which is a commonly used regression loss. Due to these two factors, we use this loss function as an example to illustrate the differences between our proposed linear normalization-based loss function and the existing two types. The total variation distance or $\ell1$-norm can be broadly formulated by the following equation:

\begin{equation}
\label{eq:tv}
 L(\textbf {p},\textbf {g}) = \sum_i|p_i-g_i|,
\end{equation}
where \textbf {p} is the predicted result and \textbf {g} is the ground-truth. The definitions of these two terms are different in each loss function, as listed in the following:

In $\ell1$-norm (unnormalized loss function),
\begin{equation}
p_i =x_i^p   ,\quad   g_i =x_i^g.
\end{equation}

In softmax normalization-based loss function,
\begin{equation}
\label{eq:softmax}
p_i =\frac{ exp(x_i^p)}{\sum_{i=1}^N exp(x_i^p)}  ,\quad   g_i =\frac{ exp(x_i^g)}{\sum_{i=1}^N exp(x_i^g)}.
\end{equation}

In linear normalization-based loss function,
\begin{equation}
\label{eq:linear}
p_i =\frac{ x_i^p}{\sum_{i=1}^N x_i^p}   ,\quad   g_i =\frac{ x_i^g}{\sum_{i=1}^N x_i^g},	
\end{equation}
where $\bm{x} = (x_1,...,x_i,...,x_N)$ is the set of unnormalized saliency response values for either the predicted saliency map (\bm{$x^p$}) and the ground-truth saliency map (\bm{$x^g$}). 

The experiments in section \ref{lossfa} illustrate that proposed linear normalization-based loss functions perform better than both softmax normalization-based and unnormalized ones. The target output in saliency prediction is an array $\bm{x^g} \in [0,1]^N$. According to the following theorem, for an array whose values between 0 and 1, the softmax will de-emphasize the maximum values among them \cite{wiki:softmax} while the linear normalization still maintains their initial proportion. Therefore, the existing loss functions coupled with softmax normalization cannot measure the gaps between the predicted probability distribution and its corresponding ground-truth very well.

\begin{theorem-non}
\label{theo:tv}	
	Given an array $\bm{x} \in [0,1]^N$, using Equation (\ref{eq:softmax}) and Equation (\ref{eq:linear}) to normalize this array separately, denote the range of the elements of this two normalized arrays as $[a_s,b_s]$ and $[a_l,b_l]$, respectively. Then, we have:
	\begin{equation*} 
					[a_s,b_s] 	\subset [a_l,b_l].
	\end{equation*}
\end{theorem-non}

\begin{proof}
It is obvious that both these normalization functions are monotonic increasing functions. We also note that $\bm{x} \in [0,1]^N$. So, we get the minimum normalized response when $x_i=0$ and get the maximum when $x_i=1$. Considering that we have $a_s = \frac{ exp(0)}{\sum_i exp(x_i)} = \frac{1}{\sum_i e^{x_i}}> 0 = \frac{0}{\sum_i x_i} = a_l$.
Now we only need to prove $b_l \geq b_s$. In fact, we have:
\begin{equation*} 
\scriptsize
b_l-b_s=\frac{ 1}{\sum_i x_i} - \frac{e}{\sum_i e^{x_i}} = \frac{ \sum_i (e^{x_i} - e x_i)  }{\sum_i x_i \sum_i e^{x_i}}. 
\end{equation*}
Recall that $x_i \in [0,1]$, it is easy to prove that $e^{x_i} - e x_i \geq 0$ for every $x_i \in [0,1]$. So we have $b_l \geq b_s$.

\end{proof}
\section{EXPERIMENTS}
\label{sec:exp}

In this section, we apply our proposed DINet for saliency prediction and report its experimental results on several public saliency benchmark datasets. The effectiveness and efficiency of our model is validated qualitatively and quantitatively. 

\subsection{Saliency Benchmark Datasets} 
For evaluating the saliency prediction model, we adopt three popular saliency benchmark datasets with different image contents and experimental settings.

\subsubsection{SALICON \cite{jiang2015salicon}}
It contains 10,000 training images, 5,000 validation images, and 5,000 testing images, taken from the Microsoft COCO dataset \cite{lin2014microsoft}. The spatial resolution of each image in this dataset is $480 \times 640$. At present, it is the largest public dataset for visual saliency prediction. The ground-truths of training and validation datasets are available while the ground-truths of test dataset are held out. For evaluation on its test dataset, researchers need to submit their results on the SALICON challenge website\footnote{https://competitions.codalab.org/competitions/3791}. Besides, the evaluation protocols and codes are available in the website\footnote{https://github.com/NUS-VIP/salicon-evaluation}.

\subsubsection{MIT1003 \cite{judd2009learning}}
It contains 1,003 images collected from Flickr and LabelMe. The ground-truths for this dataset are created from eye-tracking data of 15 users. The evaluation codes for this dataset are available in the MIT Saliency Benchmark website\footnotemark.

\subsubsection{MIT300 \cite{Judd_2012}}
It contains 300 images, including both indoor and outdoor scenarios. The ground-truths for this entire dataset are held out. Researchers can only submit the results of their models to the MIT Saliency Benchmark website\footnotemark[\value{footnote}] for evaluation. Currently, the MIT1003 dataset is usually used as the training and validation sets for this dataset.

\footnotetext{http://saliency.mit.edu/}


%
%
%

\subsection{Evaluation Metrics for Saliency Prediction} 

There exists a large variety of metrics to measure the agreement between model predictions and human eye fixations.  Following existing works \cite{salMetrics_Bylinskii,riche2013saliency}, we conduct our quantitative experiments by adopting four widely used saliency evaluation metrics, including AUC, shuffled AUC (sAUC), Normalized Scanpath Saliency (NSS), and Linear Correlation Coefficient (CC). For the sake of simplification, we denote the predicted saliency map as P, the ground-truth saliency map as G, and the ground-truth fixation map as Q. The saliency evaluation metrics are listed in Table \ref{tab:evaluation} according to their characteristics.

\begin{table}[]
	\centering
	\scriptsize
	\caption{Saliency evaluation metrics}
	\label{tab:evaluation}
	\begin{tabular}{|l|c|c|}
		\hline
		Metrics                             & Category           & Ground-truth  \\ \hline \hline
		AUC (area under the ROC curve)      & Location-based     & Fixation Map (Q)\\  \hline
		sAUC (shuffled AUC)                 & Location-based     & Fixation Map (Q)\\ \hline
		Normalized Scanpath Saliency (NSS)  & Value-based        & Fixation Map (Q)\\ \hline
		Linear Correlation Coefficient (CC) & Distribution-based & Saliency Map (G) \\ \hline
	\end{tabular}
	\vspace{-4mm}
\end{table}

\subsubsection{AUC and sAUC}
AUC means the Area Under the ROC curve. This metric evaluates the binary classification performance of the predicted saliency map P, where fixation and non-fixation points in its corresponding Q are divided into the positive set and negative set, respectively. By using a threshold, P can be binary classified into the salient and non-salient regions. ROC curve will be obtained by varying this threshold from 0 to 1. Finally, the AUC metric can be calculated by using this ROC curve. Shuffled AUC (sAUC) is introduced to alleviate the influence of center-bias. Differ in AUC, the fixation points of other images in this dataset is used as the negative set in computing sAUC values. However, these two AUC-based metrics have the limitation in penalizing false positives, as reported in \cite{kruthiventi2017deepfix,liu2016deep,cornia2016predicting}.

\subsubsection{NSS}
Normalized Scanpath Saliency (NSS) is a specific value-based saliency evaluation metric. This metric is computed by taking the mean of $\bar{P}$ at the human eye fixations $Q$:

\begin{equation}
NSS = \frac{1}{N}\sum_{i=1}^N \bar{P}(i)\times Q(i),
\end{equation}
where $N$ is the total number of human eye fixations,  $\bar{P}$ is the unit normalized saliency map $P$. 

\subsubsection{CC}
The Linear Correlation Coefficient (CC) is a statistical metric for measuring the linear correlation between two random variables. For saliency prediction evaluation, the predicted saliency maps (P) and ground-truth density maps (G) are treated as two random variables. Then, CC is calculated by the following equation:
\begin{equation}
\label{eq:loss2}
CC = \frac{cov(P,G)}{\sigma(P)\times\sigma(G)},
\end{equation}
where $cov(\cdot,\cdot) $ and $\sigma(\cdot)$ refer to the covariance and standard deviation, respectively. 




\subsection{Implementation Details} 
Our proposed DINet is implemented by Keras with TensorFlow backend \cite{chollet2015keras,abadi2016tensorflow}. During training, the weights in Dilated ResNet-50 Network (DRN) are initialized from the ResNet-50 Network (without the last fully connected layer) which is pre-trained on the ImageNet. The weights of remaining layers are initialized by the default setting of Keras. The whole model is trained with widely used Adam optimizer \cite{kingma2014adam}. A mini-batch of 10 images is used in each iteration. The learning rate is set at $10^{-4}$ and scaled down by a factor of 0.1 after every two epochs. 
We train our model on the training set of SALICON \cite{jiang2015salicon} with 10,000 training images and use its validation datasets (5,000 validation images) to validate the model. For the MIT1003 dataset \cite{judd2009learning}, we directly use the model trained on the SALICON dataset to evaluate the generalization performance of our model on this dataset. For testing on the MIT300 dataset \cite{Judd_2012}, we fine-tune our model in the MIT1003 dataset with the same evaluation protocol in \cite{cornia2016predicting,liu2016deep}. The fine-tuned results of the MIT1003 dataset are also presented. For the latter two datasets, the input images are all resized to $320 \times 480$ with zero padding to keep the original content aspect ratio. This input image size is decided by our validation experiments on SALICON dataset.

It is worth mentioning that our model can achieve processing speed as little as 0.02s and 0.03s for one input image of size $240 \times 320$ and $320 \times 480$, respectively, by using one single GTX 1080 Ti GPU. 

\subsection{Loss Function Analysis} 
\label{lossfa}
We compare the performance of our baseline models trained by our proposed probability distribution distance metrics with linear normalization to those trained on standard regression loss functions and existing softmax normalization based statistical distances. 

Table \ref{table:losscompare} presents the experimental results for each loss function, as measured by the overall performance with respect to four aforementioned evaluation metrics on SALICON validation dataset. These results support that: (i) generally, the loss functions based on probability distribution distance metrics perform better than standard regression loss functions, such as BCE, $\ell1$-norm, and $\ell2$-norm in our experiments; (ii) for a specific statistical distance based loss function, our proposed linear normalization method is more compatible than the softmax normalization as it can measure the distance between the predicted probability distribution and its target in a more proper way; (iii) Using NSS loss function alone can obtain an extremely high NSS score while this loss function is not very good at other three evaluation metrics. 

The first two conclusions have been discussed in section \ref{sec:Loss}. The reason for (iii) can be illustrated by Table \ref{tab:evaluation}. NSS is a value-based saliency evaluation metric since it is computed by the average of the normalized saliency values at eye fixation locations. In other words, a saliency map with a higher NSS score is more like a fixation map which is not similar to the fixation density map, \ie saliency map. Conversely, another three evaluation metrics (CC, AUC, sAUC) prefer the latter one. Therefore, it is difficult to use one single loss function to train the DCNN model for obtaining a promising result on both NSS and other evaluation metrics. 


\begin{table}[]
	\centering
	\scriptsize
	\caption{Performance comparison of the baseline models with different loss functions on SALICON validation dataset \cite{jiang2015salicon}.}
	\label{table:losscompare}
	\begin{tabular}{|l|c c c c|}
		\hline
		Loss function                      & CC    & sAUC  & AUC   & NSS   \\ \hline \hline
		Total Variation distance (linear)  & \textbf{0.843} & 0.788 & 0.885 & 3.077 \\ \hline
		Total Variation distance (softmax) & 0.826 & 0.786 & \textbf{0.888} & 2.906 \\ \hline
		$\ell1$-norm                & 0.810  & 0.783 & 0.874 & 2.960  \\ \hline \hline
		Bhattacharyya distance (linear)    & 0.839 & 0.786 & 0.881 & 3.077 \\ \hline
		Bhattacharyya distance (softmax)   & 0.828 & 0.785 & 0.884 & 2.992 \\ \hline \hline
		KLD (linear)             & 0.842 & 0.788 & 0.886 & 3.070  \\ \hline
		KLD (softmax)            & 0.827 & 0.785 & 0.884 & 2.968 \\ \hline\hline
		$\chi^2$ divergence (linear)             & 0.826 & \textbf{0.790}  & 0.886 & 2.994 \\ \hline
		$\chi^2$ divergence (softmax)            & 0.826 & 0.786 & 0.883 & 2.968 \\ \hline \hline
		Cosine distance (linear)           & 0.835 & 0.789 & 0.885 & 3.048 \\ \hline
		Cosine distance (softmax)          & 0.828 & 0.786 & 0.887 & 2.999 \\ \hline \hline
		BCE               & 0.826 & 0.785 & 0.885 & 2.963 \\ \hline
		Euclidean ($\ell2$-norm)                          & 0.824 & 0.781 & 0.884 & 2.958 \\ \hline\hline
		NSS                                & 0.733 & 0.782 & 0.860  & \textbf{3.411} \\ \hline
	\end{tabular}
\end{table} 

%


\subsection{Model Visualization}
 We verify the effectiveness of DINet by individually visualizing the responses of each dilated convolutional branch. This visualization experiment is realized by adding an additional decoder network without non-linear activation at the end of our DIM. Both of this additional decoder and the original decoder are jointly trained with the same loss and the same inputs from the DIM. Since the additional decoder is a linear operator applied to input feature maps, the joint decoded output in this decoder can be decoupled into a linear combination of the outputs coming from individual branches. Moreover, the input dimension of our decoder is the same as the output dimension of every branch in our DIM (all are equal to 256). The responses of each branch can be easily obtained by feeding this additional decoder with the features learned in this specific branch. By visualizing both joint and individual saliency prediction results, we can analyze the contribution of these dilated convolutional branches in our DIM.

\begin{figure}[h]
	\centering
	\scriptsize
	\includegraphics[page=6,trim = 5mm 3mm 5mm 7mm, clip,width=1.0\linewidth]{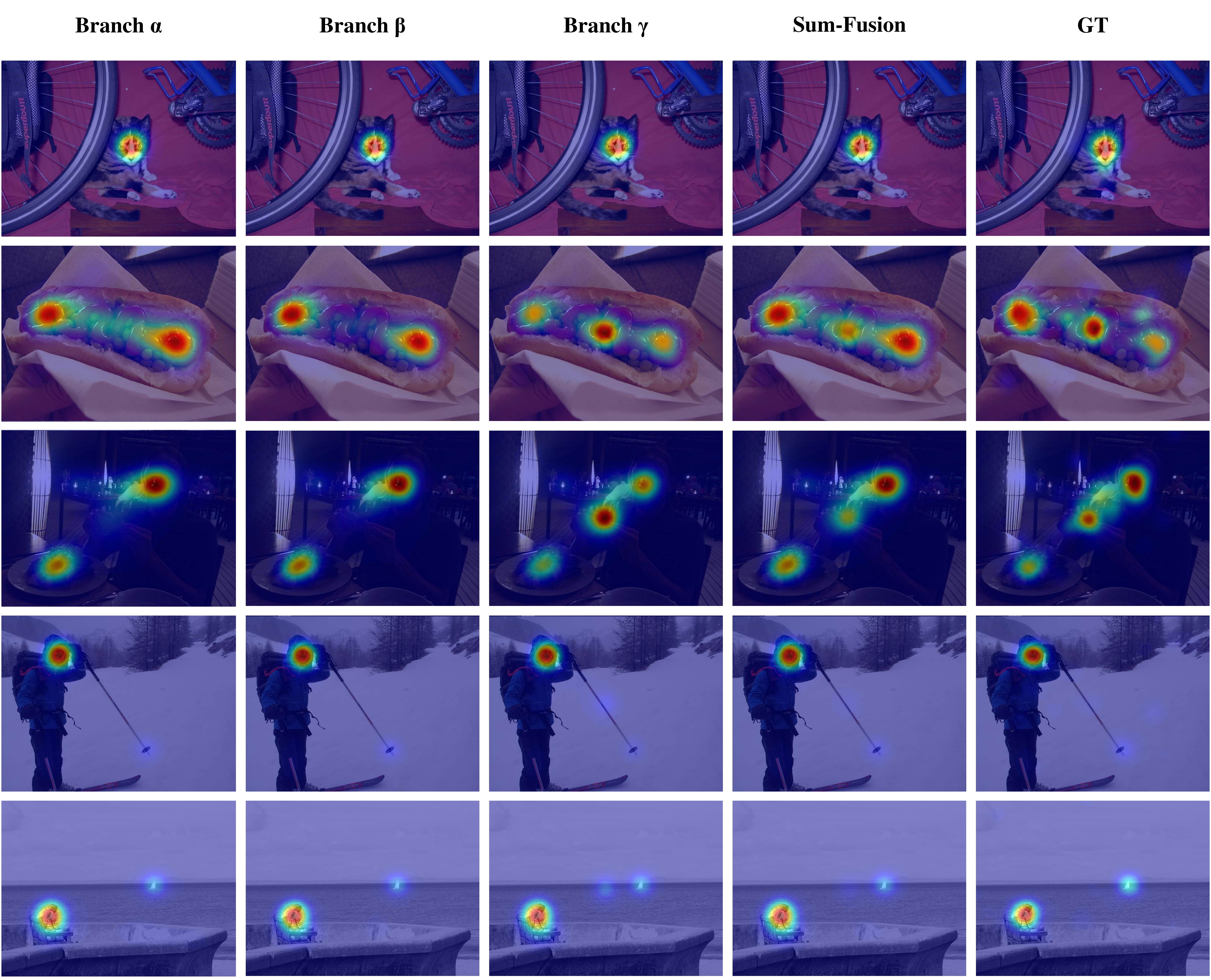}
	\caption{
The influence of each dilated convolutional branch in the DIM to visual saliency. In each col, images are the saliency prediction results by using the features captured from the above indicated branch. GT: Ground Truth.
	}
	\label{fig:inc2}
\end{figure}

	Fig. \ref{fig:inc2} demonstrates the saliency prediction results of five validation images. The first three columns show the saliency maps independently predicted by branch-$\alpha$, -$\beta$ and -$\gamma$, and the fourth column shows the final saliency maps by sum-fusing the outputs produced by mentioned branches. All of these predicted saliency maps are generally consistent with the ground-truth. As demonstrated in the second and the third rows, branches with different receptive fields learn to focus on different parts of an input image. Specifically, the branch $\gamma$, \ie $b_\gamma$, with the largest dilation rate, learns the center-bias implicitly without any additional supervision. These learned center-bias patterns compensate the negligence on the center salient regions from other two branches, $b_\alpha$ and $b_\beta$, and produce a more accurate saliency prediction result. On the other hand, $b_\gamma$ sometimes generates false alarms in the center regions with low confidence. In this case, as shown in the last two rows of Fig. \ref{fig:inc2}, the previous two branches $b_\alpha$ and $b_\beta$ can help in reducing this unwilling side-effect on the final fusion results. These three branches in our DIM work in a collaborative manner. The results by using the features from a single branch are no need to be perfect for all possible cases. These incomplete predictions will be ensembled by the sum-fusion to become more comprehensive and reliable final results, which can be also supported by our ablation analysis in Table \ref{table:dimaa1}.

%
%

%

\subsection{Model Ablation Analysis} 
In this section, we conduct ablation analysis for our DINet on the SALICON validation dataset. The complete ablation results are presented in Table \ref{table:maa}. It should be noted that all of models in this table are trained by the proposed linear normalization-based total variation distance loss function.

\begin{table}[]
	\centering
	\scriptsize
	\caption{
		Model ablation analysis on SALICON validation dataset \cite{jiang2015salicon}.
		}
	\label{table:maa}
	\begin{tabular}{|l|cccc|}
		\hline
		Model                                & CC     & sAUC  & AUC   & NSS   \\ \hline\hline
		\multicolumn{5}{|c|}{Influence of backbone network}               \\ \hline
		ResNet + Decoder               & 0.776  & 0.762  & 0.879 & 2.456 \\ \hline
		Baseline (DRN + Decoder)       & 0.843  & 0.788 & 0.885 & 3.077 \\ \hline\hline
		\multicolumn{5}{|c|}{Influence of decoder network}               \\ \hline
		DRN + Decoder(1 conv layer)               & 0.838  & 0.785  & 0.883 & 3.052 \\ \hline
		DRN + Decoder(2 conv layers)               & 0.841  & 0.787 & 0.884 & 3.067\\ \hline
		DRN + Decoder(4 conv layers)               & 0.843  & 0.788  & 0.885 & 3.072 \\ \hline
		DRN + Decoder(1 deconv + 1 conv layers)       &0.841   & 0.787 & 0.884  & 3.064 \\ \hline			
		DRN + Decoder(2 deconv + 1 conv layers)       & 0.841  & 0.788 & 0.884 & 3.067 \\ \hline
		DRN + Decoder(3 deconv + 1 conv layers)       & 0.841  & 0.787  & 0.885 & 3.061 \\ \hline\hline						
		\multicolumn{5}{|c|}{Effectiveness of multi-scale features} \\ \hline
		ResNet + Skip-layer + Decoder & 0.841  & 0.786  & 0.885 & 3.053 \\ \hline 
		Baseline + IPN                       & 0.849   & 0.787 & 0.885 & 3.086 \\ \hline
		Baseline + Skip-layer                      & 0.847  & 0.788  & 0.886 & 3.084 \\ \hline
		Baseline + Inception(a)            & 0.850   & 0.788 & 0.886 & 3.094 \\ \hline
		Baseline + Inception(a) - 1$\times $1 branch      &0.849	&0.789	&0.886	&3.091 \\ \hline
		Baseline + Inception(b)             &0.852	&0.790	&0.886	&3.107 \\ \hline		
		Baseline + Inception(c)             & 0.852	&0.790	&0.886	&3.111 \\ \hline		
		Baseline + Inception(d)             & 0.854	&0.790	&0.887	&3.114 \\ \hline
		DINet (Baseline + Inception(e))  & 0.853  & 0.789 & 0.887 & 3.117 \\ \hline
		DRN + ASPP-S             & 0.853&	0.789&	0.887	&3.112 \\ \hline		
		DRN + ASPP-L            & 0.852&	0.789	&0.887	&3.102 \\ \hline \hline			
		\multicolumn{5}{|c|}{Influence of training image size}               \\ \hline
		DINet ($240 \times 320$)                      & 0.853  & 0.789 & 0.887 & 3.117 \\ \hline
		DINet ($320 \times 480$)                      & 0.858  & 0.790  & 0.887 & 3.143 \\ \hline
		DINet ($480 \times 640$)                      & 0.854  & 0.789 & 0.886 & 3.128 \\ \hline
		DINet    (ensemble)                  & \textbf{0.867}  & \textbf{0.792}  & \textbf{0.889} & \textbf{3.168} \\ \hline
	\end{tabular}
\vspace{-4mm}
\end{table}

\subsubsection{Influence of the backbone network} 
Our baseline model is built on DRN where the \emph{output\_stride} is equal to 8. As mentioned in Section \ref{sec:related}, the \emph{output\_stride} of original ResNet is 32 which means that less spatial information are included in the output of this backbone network and thus leads to the unsatisfactory performance. To verify this statement, we compare our baseline model (DRN + decoder) with a more basic model (ResNet + decoder). From the first part of Table \ref{table:maa}, we can conclude that \emph{output\_stride} is one of the key elements for the dense prediction tasks. There is a significant performance gain by replacing the original ResNet with DRN.

\subsubsection{Influence of the decoder network} 
\label{sec:dec}
In our baseline model, our designed decoder network is just three convolutional layers plus sigmoid activation in the end. The reason for using three layers is determined by the experiments. We have tried to use different number of convolutional or deconvolutional layers before the prediction layer (one convolutional layer followed by a sigmoid activation) to form other decoder networks. Their results are reported in the second part of Table \ref{table:maa}. As we can see that the models with these decoders cannot get good results as our original decoder, \ie Decoder(3 conv layers).

%

\subsubsection{Effectiveness of multi-scale features} 

DINet uses the proposed DIM to capture multi-scale contextual features. To support the conclusions in \cite{huang2015salicon,kruthiventi2017deepfix,liu2016deep} that integrating multi-scale features can further improve saliency detection performance, we incorporate existing alternative multi-scale feature extraction modules, including IPN, skip-layer, inception and ASPP, into our baseline or backbone network. From the third part of Table \ref{table:maa}, we can observe that the saliency prediction performance indeed boosted by incorporating the multi-scale features. Especially, when the backbone network is not DRN, the multi-scale features can compensate the performance drop significantly, by comparing two models with the plain ResNet backbone network. In all these multi-scale saliency prediction framework, our proposed inception(d) and (e) obtain the optimal results among them. For the reason that inception(e) is more efficient in terms of \#parameters and inference time, as illustrated by Table \ref{inception1}, we pick this dilated inception module to form our DINet.

\subsubsection{Ablation analysis on DIM} 
 We further verify the effectiveness of our DIM by conducting two quantitative experiments. In the first experiment, we evaluate the performance of a trained DINet with two decoders mentioned in the visualization experiment to investigate the contribution of each dilated convolutional branch in our DIM respectively. In the second experiment, we make a comparison among a set of variants of DINet to explore the impact of the number of parallel dilated convolutional layers.

%


\begin{table}[]
	\centering
	\scriptsize
	\caption{
Dilated inception module ablation analysis within a trained DINet with two decoders on SALICON validation dataset \cite{jiang2015salicon}.
	}
	\label{table:dimaa1}
	\begin{tabular}{|l|ccc|cccc|}
		\hline
		Type                          &$b_\alpha$ & $b_\beta$ &$b_\gamma$ & CC    & sAUC  & AUC   & NSS   \\ \hline \hline
		\multicolumn{8}{|c|}{The results on additional decoder network}               \\ \hline
		0 branch                 &    &    &    &  0.752  & 0.794 & 0.858 & 2.729 \\ \hline
		\multirow{3}{*}{1 branch}     & \checkmark  &    &    & 0.833 & 0.799 & 0.882 & 3.012 \\ \cline{2-8} 
		&    & \checkmark  &    & 0.811  & 0.793 & 0.864 & 3.025 \\ \cline{2-8} 
		&    &    & \checkmark  & 0.801 & 0.759  & 0.873 & 2.967  \\ \hline
		\multirow{3}{*}{2 branches-sum} & \checkmark  & \checkmark  &    & 0.831 & 0.799 & 0.879  & 3.035 \\ \cline{2-8} 
		& \checkmark  &    & \checkmark  & 0.804 &0.799 & 0.869 & 3.036 \\ \cline{2-8} 
		&    & \checkmark  & \checkmark  & 0.813 &\textbf{0.800} & 0.872 & 3.032 \\ \hline
		3 branches-sum                 & \checkmark  & \checkmark  & \checkmark  & \textbf{0.853} & 0.789 & 0.886 & 3.098 \\ \hline \hline
		\multicolumn{8}{|c|}{The results on original decoder network}               \\ \hline
		3 branches-sum                 & \checkmark  & \checkmark  & \checkmark  & \textbf{0.853} & 0.789 & \textbf{0.887} & \textbf{3.107} \\ \hline		
	\end{tabular}
\vspace{-4mm}
\end{table}

\begin{table}[]
	\centering
	\scriptsize
	
	\caption{
Dilated inception module ablation analysis with individual trained variants of DINet on SALICON validation dataset \cite{jiang2015salicon}.
	} 
	\label{table:dimaa2}
	\begin{tabular}{|l|ccc|cccc|}
		\hline
		Type                            & $b_\alpha$ & $b_\beta$ &$b_\gamma$ & CC    & sAUC  & AUC   & NSS   \\ \hline
		0 branch                &    &    &    &  0.843  & 0.788 & 0.885 & 3.077 \\ \hline
		
		\multirow{3}{*}{1 branch}       & \checkmark   &    &    & 0.847 & \textbf{0.790}  & 0.886 & 3.080  \\ \cline{2-8} 
		&    & \checkmark   &    & 0.849 & 0.788 & \textbf{0.887} & 3.086 \\ \cline{2-8} 
		&    &    & \checkmark   & 0.851 & 0.788 & \textbf{0.887} & 3.095 \\ \hline
		\multirow{3}{*}{2 branches-sum} & \checkmark   & \checkmark   &    & 0.852 & 0.788 & \textbf{0.887} & 3.099 \\ \cline{2-8} 
		& \checkmark   &    & \checkmark   & 0.853 & 0.788 & 0.886 & 3.098 \\ \cline{2-8} 
		&    & \checkmark   & \checkmark   & 0.852 & 0.788 & \textbf{0.887} & 3.103 \\ \hline
		3 branches-sum                  & \checkmark   & \checkmark   & \checkmark   & 0.853 & 0.789 & \textbf{0.887} & \textbf{3.117} \\ \hline
		3 branches-concat               & \checkmark   & \checkmark   & \checkmark   & \textbf{0.854} & 0.789 & \textbf{0.887} & 3.116 \\ \hline 
	\end{tabular}
\vspace{-4mm}
\end{table}


Table \ref{table:dimaa1} shows the results of the first experiment. Each row in this table represents the evaluation results by using the outputs from the indicated branch(es) as the input to a trained decoder. As we can see that, 1 branch type of DIM will learn different bias under its specific receptive fields to help in predicting visual saliency. Specifically, $b_\alpha$ prefers the results with higher sAUC score, while $b_\beta$ is more interested in the NSS metric. By comparing the results between the row of 3 branches-sum and the rows in 2 branches-sum type on the first part of this table, we can observe that the performance drop dramatically with the absence of any one branch, which means every branch in our DIM has its irreplaceable impact on the final results. These three branches in our DIM work in a collaborative manner. Even if the performance by using any individual branch is not comparable to the performance of our baseline model, their fused results can deal with the diverse images with different patterns of salient regions. Moreover, the results on the last row show that the features used in the additional decoder can still be decoded by our original decoder with only a little bit performance drop in the NSS metric. It can guarantee the generality of the above conclusions.

Table \ref{table:dimaa2} compares the performance of several variants of DINet.
 Each row in this table means the evaluation results by testing the individual trained variant which has the indicated branch(es). Especially, the model in 3 branches-sum type is the proposed DINet, while the model in 0 branch type is our baseline model. This table shows that using more branches (from 0 to 3), which means using more comprehensive features, will lead to a higher performance on evaluation metrics. Besides, in the 1 branch type of DINet, using dilated convolution with larger dilation rate before the decoder network can achieve a better performance than using a smaller one. It can be credited to the larger size of receptive fields which represent the longer range of dependencies in captured features. Moreover, using concatenation to replace our element-wise addition has a limited impact on the final results, as presented in the last two rows in this table. Mathematically, element-wise addition followed by a convolution layer is a special case of concatenation followed by another convolution layer \cite{chen2017dual}, which can be used to explain this limited difference on evaluation results. In summary, both of these two experiments can verify that the performance gain of our DIM is realized by the corporation of these three parallel dilated convolutional branches.

\subsubsection{Influence of training image size} 
The previous experimental results on SALICON validation dataset are all obtained from $240 \times 320$ images, whose size is the half resolution of the original SALICON images. Here we want to see the performance of our DINet models which are trained by images with different spatial resolution. From Table \ref{table:maa}, we find that the DINet trained by input images of size $320 \times 480$ can obtain the best performance among these three models. This model will be directly fine-tuned in the MIT1003 dataset for the evaluation of the MIT300 dataset. Note that these evaluation results are the average scores, there are some validation images which perform better in other DINets ($240 \times 320$ or $480 \times 640$). In order to characterize this phenomenon, we adopt a simple ensemble learning metric, i.e. average voting, to further improve the performance of our model. By using the average results from these three different models, this ensemble model obtain the best scores in our model ablation analysis.

\subsubsection{Ensemble learning for improving NSS} 
However, our best model, which is trained by a single total variation distance loss function, still cannot beat two state-of-the-art models \cite{cornia2016predicting,liu2016deep} in NSS metrics, as shown in Table \ref{table:lossen}. These two models use the NSS itself as one of the loss functions for training. To further improve our performance on NSS metrics, we use the same ensemble learning method as above to combine the results of two DINet models which are trained by using two different loss function (total variation distance with linear normalization and NSS) separately. The last ensemble model in this table is our final submission to the SALICON test dataset which results in a good comprise between NSS and another three evaluation metrics.

\begin{table}[]
	\centering
	\scriptsize
	\caption{Performance comparison of our DINet models with different loss functions on SALICON validation dataset \cite{jiang2015salicon}.}
	\label{table:lossen}
	\begin{tabular}{|l|cccc|}
		\hline
		Model                                & CC    & sAUC  & AUC   & NSS   \\ \hline 
		DINet (TV distance)                  & \textbf{0.867} & \textbf{0.792} & \textbf{0.889} & 3.168 \\ \hline 
		DSCLSTM     \cite{liu2016deep}                         & 0.835 & 0.788 & 0.887 & 3.221 \\ \hline
		SAM-ResNet  \cite{cornia2016predicting}                         & 0.844 & 0.787 & 0.886 & 3.260  \\ \hline 
		DINet (NSS)                           & 0.724 & 0.782 & 0.861 & \textbf{3.600}   \\ \hline 
		DINet (ensemble NSS and TV distance)) & 0.862 & \textbf{0.792} & 0.886 & 3.310  \\ \hline 
	\end{tabular}
\vspace{-4mm}
\end{table}

\subsection{Comparison with state-of-the-arts}

To demonstrate the effectiveness of our proposed DINet model in predicting visual saliency, we quantitatively compare our method with state-of-the-art models on SALICON, MIT1003, and MIT300 datasets. 


\begin{table}[]
	\scriptsize
	\centering
	\caption{Comparison results on the SALICON test dataset \cite{jiang2015salicon}.}

	\label{salicontest}
	\begin{tabular}{|l|c c c c|}
		\hline
		Models & CC & sAUC & AUC & NSS \\ \hline  
		\textbf{DINet (Ours)} & \textbf{0.860} & 0.782 & \textbf{0.884} & \textbf{3.249} \\ \hline
		SAM-ResNet \cite{cornia2016predicting} & 0.842 & 0.779 & 0.883 & 3.204\\ \hline
		DSCLRCN \cite{liu2016deep} & 0.831 & 0.776 & \textbf{0.884} & 3.157 \\ \hline
		SAM-VGG \cite{cornia2016predicting} & 0.825 & 0.774 & 0.881 & 3.143 \\ \hline
		SalGAN \cite{pan2017salgan} & 0.781 & 0.772 & 0.781 & 2.459 \\ \hline
		SU \cite{kruthiventi2016saliency} & 0.780 & 0.760 & 0.880 & 2.610 \\ \hline
		PDP \cite{jetley2016end}& 0.765 & 0.781 & 0.882 & - \\ \hline
		ML-Net \cite{cornia2016deep} & 0.743 & 0.768 & 0.866 & 2.789 \\ \hline
		MxSalNet \cite{dodge2017visual} & 0.730 & 0.771 & 0.861 & 2.767 \\ \hline
		Deep Convnet \cite{pan2016shallow} & 0.622 & 0.724 & 0.858 & 1.859 \\ \hline
		Shallow Convnet \cite{pan2016shallow} & 0.562 & 0.658 & 0.821 & 1.663 \\ \hline
		DeepGazeII \cite{kummerer2016deepgaze} & 0.479 & \textbf{0.787} & 0.867 & 1.271 \\ \hline		
	\end{tabular}
	\vspace{-4mm}
\end{table}

\begin{figure*}[h]
	%
	%
	
	\centering
	\includegraphics[page=5,trim = 5mm 4mm 5mm 5mm, clip, width=1.0\linewidth]{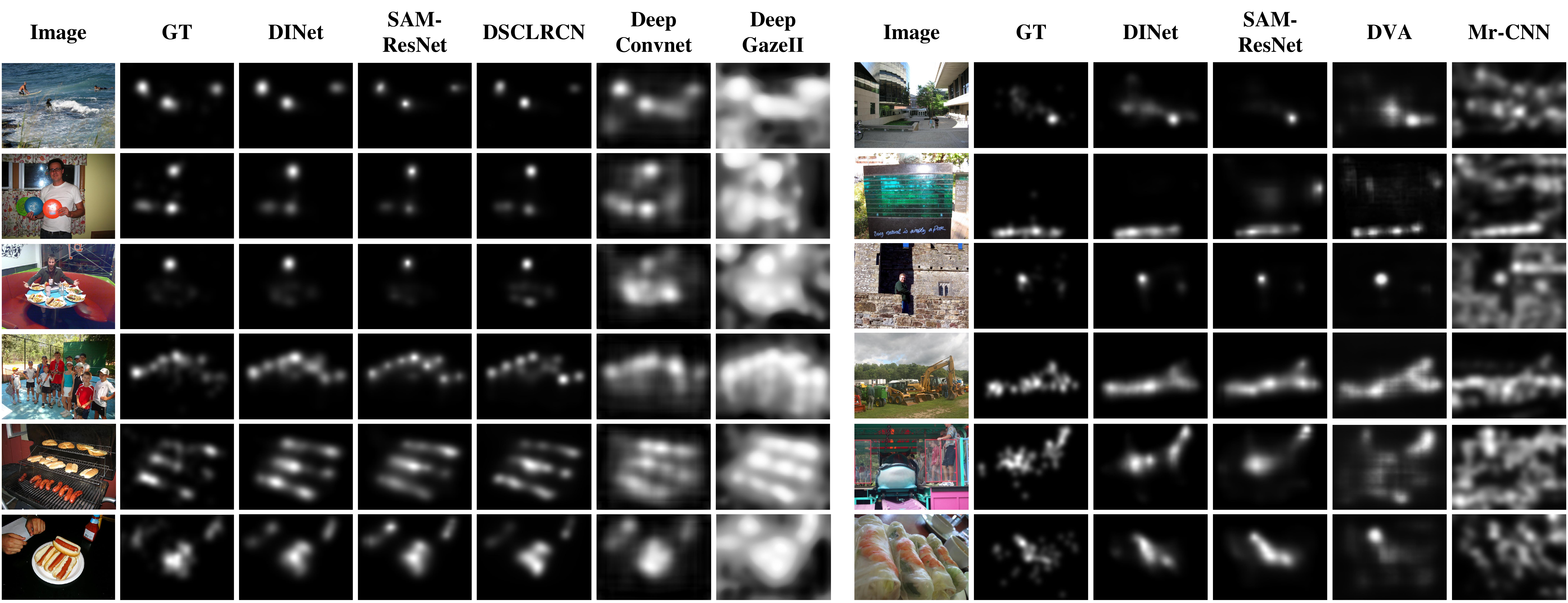}
	\caption{Qualitative comparison results on two datasets. Left images are from SALICON validation dataset \cite{jiang2015salicon}, while right images are from MIT1003 dataset \cite{judd2009learning}. GT: Ground Truth.}
	\label{fig:salicon}
\end{figure*}

Table \ref{salicontest} shows the evaluation results on the SALICON dataset. The results of other models come from their papers or the leaderboard of this dataset. In this table, the results in bold indicate the best performance method on each evaluation metric. As it can be observed, our DINet outperforms all competitors on CC, AUC, and NSS three metrics. The DeepGazeII \cite{kummerer2016deepgaze} model get the best sAUC score and relatively lower scores on other metrics. The saliency maps generated by this model actually are very blurred/hazy and visually different from the ground-truth, as shown in the left part of Fig.\ref{fig:salicon}. This is because AUC-based metrics mainly relied on true positives without significantly penalizing false positives \cite{kruthiventi2017deepfix,cornia2016predicting}. 




\begin{table}[]
	\centering
	\caption{Comparison results on the MIT1003 dataset \cite{judd2009learning}.}
	\label{MIT1003}
	\begin{tabular}{|l|c c c c|}
		\hline
		Model                                     & CC   & sAUC & AUC  & NSS  \\ \hline
		\textbf{DINet (w/o finetune)}                      & \textbf{0.67} & 0.70  &\textbf{0.88} &\textbf{2.40}  \\ \hline
		DVA \cite{wang2017deep}                                      & 0.64 & \textbf{0.77} & 0.87 & 2.38  \\ \hline
		GBVS    \cite{harel2007graph}                                  & 0.42 & 0.66 & 0.83 & 1.38  \\ \hline
		eDN    \cite{vig2014large}                                   & 0.41 & 0.66 & 0.85 & 1.29   \\ \hline		
		Mr-CNN \cite{liu2016learning}                                   & 0.38 & 0.73 & 0.80  & 1.36  \\ \hline
		BMS    \cite{zhang2013saliency}                                    & 0.36 & 0.69 & 0.79 & 1.25  \\ \hline
		ITTI   \cite{itti1998model}                                    & 0.33 & 0.66 & 0.77 & 1.10  \\ \hline
	\end{tabular}
	\vspace{-4mm}
\end{table}

\begin{table}[]
	\centering
	\caption{Comparison results on the MIT1003 validation dataset \cite{judd2009learning}. }
	\label{MIT10031}
	\begin{tabular}{|l|c c c c|}
		\hline
		Model               & CC   & sAUC & AUC  & NSS   \\ \hline \hline
		\textbf{DINet (w finetune)}   & \textbf{0.87} & \textbf{0.77} & \textbf{0.91} & \textbf{3.27} \\ \hline
		SAM-ResNet  \cite{cornia2016predicting}               & 0.77 & 0.62 & \textbf{0.91} & 2.89   \\ \hline
		SAM-VGG     \cite{cornia2016predicting}           & 0.76 & 0.61 & \textbf{0.91} & 2.85   \\ \hline		
		DeepFix     \cite{kruthiventi2017deepfix}        & 0.72 & 0.74 & 0.90  & 2.58  \\ \hline	\hline	
		\textbf{DINet (w/o finetune)} & 0.67 & 0.72 & 0.89 & 2.50  \\ \hline
	\end{tabular}
	\vspace{-4mm}
\end{table}

\begin{table}[]
	\centering
	\caption{Comparison results on the MIT300 dataset \cite{Judd_2012}. }
	\label{mit300}
	\begin{tabular}{|l|c c c c |}
		\hline
		Model      & CC   & sAUC & AUC  & NSS   \\ \hline
		DSCLRCN \cite{liu2016deep}   & \textbf{0.80}  & 0.72 & 0.87 & \textbf{2.35}  \\ \hline
		\textbf{DINet (Ours)}      & 0.79 & 0.71 & 0.86 & 2.33  \\ \hline
		SAM-ResNet \cite{cornia2016predicting}   & 0.78 & 0.70  & 0.87 & 2.34  \\ \hline
		DeepFix \cite{kruthiventi2017deepfix}       & 0.78 & 0.71 & 0.87 & 2.26  \\ \hline
		SAM-VGG \cite{cornia2016predicting}      & 0.77 & 0.71 & 0.87 & 2.30  \\ \hline
		SALICON \cite{huang2015salicon}    & 0.74 & \textbf{0.74} & 0.87 & 2.12  \\ \hline
		SalGAN \cite{pan2017salgan}     & 0.73 & 0.72 & 0.86 & 2.04 \\ \hline
		PDP   \cite{jetley2016end}      & 0.70  & 0.73 & 0.85 & 2.05   \\ \hline
		DVA   \cite{wang2017deep}           & 0.68 & 0.71 & 0.85 & 1.98  \\ \hline
		ML-Net  \cite{cornia2016deep}    & 0.67 & 0.70  & 0.85 & 2.05  \\ \hline
				SalNet  \cite{pan2016shallow}    & 0.58 & 0.69 & 0.83 & 1.51 \\ \hline
		BMS     \cite{zhang2013saliency}    & 0.55 & 0.65 & 0.83 & 1.41 \\ \hline
		DeepGazeII  \cite{kummerer2016deepgaze} & 0.52 & 0.72 & \textbf{0.88} & 1.29 \\ \hline
		GBVS     \cite{harel2007graph}      & 0.48 & 0.63 & 0.81 & 1.24 \\ \hline
		Mr-CNN  \cite{liu2016learning}       & 0.48 & 0.69 & 0.79 & 1.37 \\ \hline
		eDN     \cite{vig2014large}     & 0.45 & 0.62 & 0.82 & 1.14 \\ \hline
		ITTI   \cite{itti1998model}      & 0.37 & 0.63 & 0.75 & 0.97  \\ \hline
	\end{tabular}
	\vspace{-4mm}
\end{table}

 \begin{table*}[]
	\centering
	\scriptsize
	\begin{threeparttable}
	\caption{
		comprehensive comparison with the state-of-the-arts.
		}
	\label{stoa}
	\begin{tabular}{|l|c|c|c|c|c|}
		\hline
		Model                  & Backbone network                  & \#parameters              & key module                                                                 & Input image size  & inference time (per image) \\ \hline
		\multirow{3}{*}{DINet (Ours)} & \multirow{3}{*}{Dilated ResNet50} & \multirow{3}{*}{27.04M} & \multirowcell{3}{Dilated inception module for multi-scale} & $240 \times 320$           & 0.02s          \\ \cline{5-6} 
		&                                   &                         &                                                                             & $320 \times 480$        & 0.03s          \\ \cline{5-6} 
		&                                   &                         &                                                                             & $480 \times 640$           & 0.06s          \\ \hline
		\multirow{2}{*}{SAM$^a$  \cite{cornia2016predicting}}   & Dilated ResNet50                  & 70.09M                  & \multirow{2}{*}{Conv-LSTMs \cite{xingjian2015convolutional} for iterative refinement}                        & \multirow{2}{*}{$240 \times 320$}           & 0.09s        \\ \cline{2-3} \cline{6-6} 
		& Dilated VGG16                     & 51.84M                  &                                                                             &            & 0.07s         \\ \hline
		DSCLRCN$^b$   \cite{liu2016deep}              & Dilated ResNet50 + Places-CNN     & \textgreater{}33.71M    & Spatial LSTMs \cite{visin2015renet} for context incorporation                                        & $480 \times 640+227 \times 227$  & 0.27s         \\ \hline
		DVA$^c$   \cite{wang2017deep}                   & VGG16                             & 25.07M                  & Skip-layers for multi-scale                               & $224 \times 224$         & 0.02s          \\ \hline
	\end{tabular}
    \begin{tablenotes}
	\small
	\item 	$a$: The codes for these models are from the authors' github website. We test its inference time in our experimental environment.
	\item  $b$: This model has many customized operations which is hard to count their trainable parameters completely and reimplement in the Keras framework. 0.27s is adopted from their paper. It is certain that this model need more parameters and longer inference time than our method.
	\item  $c$: This model is reimplemented by ours and tested in our experimental environment. Its deconvolutions-based decoder network slow down the whole model and thus it has similar inference time as our model. 
\end{tablenotes}
\end{threeparttable}
	\vspace{-4mm}

\end{table*}

The results on MIT1003 are reported in Table \ref{MIT1003}. We directly use the DINet trained on the SALICON dataset to evaluate the generalization performance of our model on the whole MIT1003 dataset, as the DVA model \cite{wang2017deep}. Our model also achieves promising results on this dataset which verifies its robustness and generality. Qualitative comparison results of our model with other state-of-the-art saliency models on SALICON validation and MIT1003 datasets can be found in Fig.\ref{fig:salicon}. This figure can also support that our results match the ground-truth saliency maps best among all the compared models in both two datasets.

In order to evaluate the MIT300 dataset, we fine-tune our DINet on the MIT1003 dataset. The fine-tuned results are shown in Table \ref{MIT10031}, As we can see that, the performance of our model improves significantly after fine-tuning which can also outperform other existing fine-tuned models. The results on MIT300 dataset are presented in Table \ref{mit300}. Different in the previous two datasets, our DINet can not outperform the DSCLRCN model \cite{liu2016deep}. Our model may over-fitted on the MIT1003 dataset which leads to lower generalization performance on MIT300 dataset. Both DSCLRCN model and our DINet use multi-scale features to further improve saliency prediction performance. Besides, DSCLRCN model incorporates the global context and scene context by using spatial LSTM \cite{visin2015renet} method and additional Places-CNN \cite{zhou2014learning} backbone network to achieve this performance. Consequently, their model is more complex and much slower than our method. When testing one image with size $480 \times 640$, the DSCLRCN model needs 0.27s while our DINet needs only 0.06s.

A comprehensive summary of our model and other three state-of-the-art competitors, \ie SAM \cite{cornia2016predicting}, DSCLRCN \cite{liu2016deep}, and DVA \cite{wang2017deep}, are listed in Table \ref{stoa}. Apart from achieving superior performance on the SALICON and MIT1003 datasets, our model also has the obvious advantages in terms of both \#parameters and inference time compared to the SAM and DSCLRCN models. 

\begin{figure}[h]
	\centering
	\scriptsize
	\includegraphics[page=7,trim = 5mm 3mm 5mm 7mm, clip,width=1.0\linewidth]{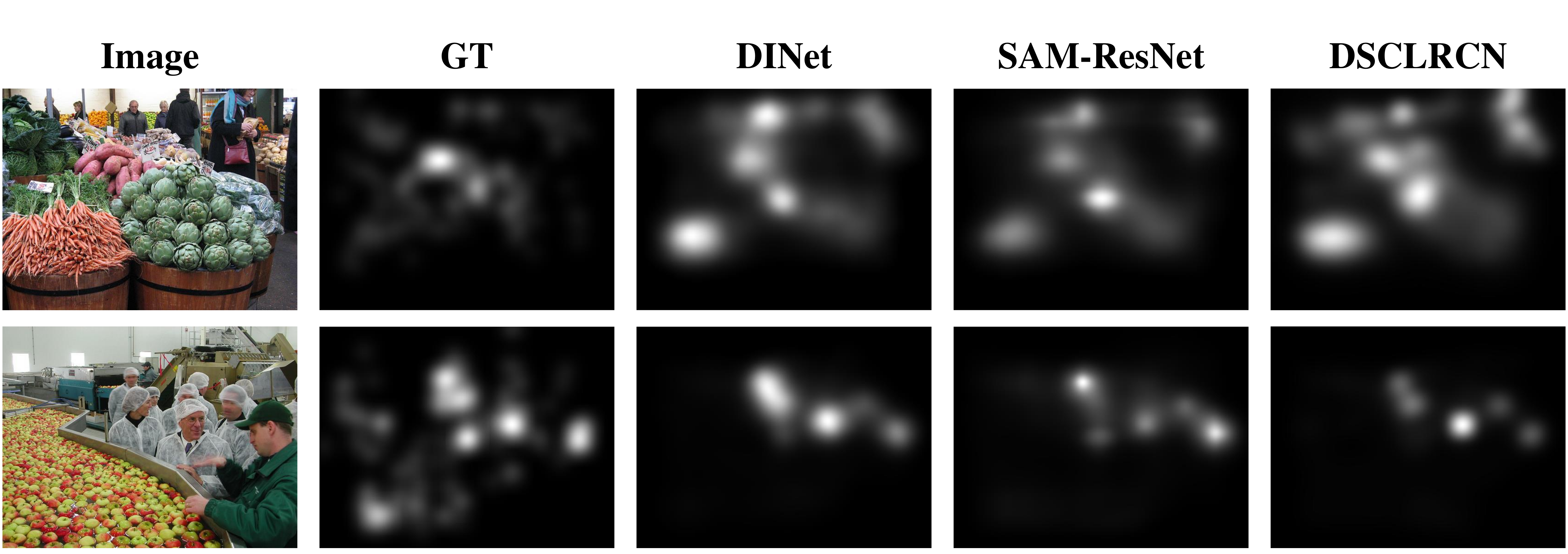}
	\caption{ Some failure cases of our DINet and two state-of-the-arts. Images are from SALICON validation dataset \cite{jiang2015salicon}.
	}
	\label{fig:fc1}
	\vspace{-4mm}
\end{figure}

However, despite the good results, there are still a small number of failure cases, as shown in Fig. \ref{fig:fc1}. These bad cases are caused by the fact that so many objects are cumulated in a single image. Within them, the relative importance of these objects cannot be fully learned by simply utilizing the multi-scale contextual features without higher level visual understanding. Therefore, some non-salient regions are highlighted (like the first row) or some salient regions are missed, as shown in the second row. Note that SAM and DSCLRCN models suffer from the same problem as ours. It can be concluded that even the state-of-the-art saliency models still cannot fully understand the relative importance of image regions in such semantically rich scenes. To further approach human-level performance, saliency models will need to discover increasingly higher-level concepts in images for determining an appropriate amount of visual attention on a certain image region.

\section{Conclusion}
\label{sec:con}
We have proposed a dilated inception network for visual saliency prediction. The multi-scale saliency-influential factors are captured by an efficient and effective dilated inception module (DIM). The whole model works in a fully convolutional encoder-decoder architecture. It is trained end-to-end and lightweight for time-efficiency. Furthermore, we adopted a set of linear normalization-based probability distribution distance metrics as loss functions to formulate the saliency prediction problem as a probability distribution prediction task. With such loss functions, our models can perform better than those trained by using either standard regression loss functions or existing softmax normalization-based probability distribution distance metrics. Experimental results on the challenging saliency benchmark datasets have demonstrated the outstanding performance of our model with respect to other relevant saliency prediction methods.

\bibliographystyle{IEEEtran}
\balance
\bibliography{bib/deep_learning_tutorial,bib/ref_VA,bib/ref_VADL,bib/ref_SOD}

%
%
%

\end{document}